\documentclass[sigconf,nonacm]{acmart}

\usepackage{microtype}
\usepackage{graphicx}
\usepackage{amsmath}
\usepackage{booktabs}

\title{When Should Active RAG Retrieve? A Budget-Aware Evaluation of Utility, Calibration, and Cost}

\author{Pin Qian}
\email{pqian@alumni.cmu.edu}
\affiliation{%
  \institution{Carnegie Mellon University}
  \city{Pittsburgh}
  \state{PA}
  \country{USA}}

\author{Su Wang}
\email{suwang@alumni.cmu.edu}
\affiliation{%
  \institution{Carnegie Mellon University}
  \city{Pittsburgh}
  \state{PA}
  \country{USA}}

\author{Chong Peng}
\email{chongp@alumni.cmu.edu}
\affiliation{%
  \institution{Carnegie Mellon University}
  \city{Pittsburgh}
  \state{PA}
  \country{USA}}

\author{Junxian You}
\email{3163509Y@student.gla.ac.uk}
\affiliation{%
  \institution{University of Glasgow}
  \city{Glasgow}
  \country{United Kingdom}}

\author{Lifei Liu}
\email{lliu.lifei@gmail.com}
\affiliation{%
  \institution{Independent Researcher}
  \city{Seattle}
  \state{WA}
  \country{USA}}

\author{Haoran Yu}
\email{haoranyu889@gmail.com}
\affiliation{%
  \institution{Independent Researcher}
  \city{Seattle}
  \state{WA}
  \country{USA}}

\author{Yihang Chen}
\email{ychen3726@gatech.edu}
\affiliation{%
  \institution{Georgia Institute of Technology}
  \city{Atlanta}
  \state{GA}
  \country{USA}}

\author{Xiaochong Jiang}
\email{jiang.xiaoc@northeastern.edu}
\affiliation{%
  \institution{Independent Researcher}
  \city{Seattle}
  \state{WA}
  \country{USA}}

\begin{document}

\begin{abstract}
Active RAG systems decide when to retrieve external knowledge during generation,
making them a budget-sensitive case of agentic RAG and self-adaptive retrieval.  Yet
evaluations often leave the operating point underspecified: two systems may
both claim a 50\% evidence-usage budget while realizing different held-out
usage rates, so higher accuracy can reflect a looser budget rather than a
better retrieval policy.  We study budget-aware evaluation for Active RAG by
recasting active retrieval as utility estimation, where retrieval is valuable
only through its marginal correctness change over a no-retrieval answer.  This
view separates three questions that single-point evaluations conflate: whether
trigger scores rank useful retrieval decisions, whether thresholds calibrated
on past data meet future budgets, and how trigger-side computation changes
deployment cost.  We operationalize these questions with exact top-$k$ utility
frontiers, deployable threshold frontiers, conservative budget frontiers, harm
audits, and cost decompositions.  Across knowledge-intensive multi-hop QA datasets and open
instruction models, retrieval harm is non-negligible, router rankings change
across datasets and budgets, nominal thresholds can miss target usage, and
simple uncertainty or retrieval-score baselines often rival learned utility
routers.  Budget-aware Active RAG evaluations should therefore report
frontiers, realized usage, threshold-transfer error, harm rates, and cost
decompositions alongside accuracy.
\end{abstract}



\maketitle

\section{Introduction}

Knowledgeable foundation models increasingly need to decide when to rely on
parametric memory and when to use retrieved evidence.  Active RAG methods
respond by asking when a model should retrieve or use evidence
\citep{jiang2023flare,asai2024selfrag,jeong2024adaptiverag,
su2024dragin,cheng2024uar}.  In practice, however, the phrase ``when to
retrieve'' hides an operating-point problem.  If two systems are both evaluated
as adaptive RAG pipelines, but one uses retrieved evidence more often, an
accuracy comparison alone does not tell us which system made better retrieval
decisions.

The deployment question is therefore sharper: under a fixed evidence-usage or
compute budget, which inputs deserve retrieval, and will the threshold chosen on
a calibration split respect that budget on future inputs?  This distinction is
not cosmetic.  In our 2,000-example Qwen2.5-1.5B experiments, routers calibrated
to a nominal 50\% evidence-usage target realize different held-out usage rates
across random splits.  On HotpotQA, common triggers exceed the target in
40\%--100\% of splits when any overshoot is counted.  The overshoot magnitude
is not always large, so we also report tolerance-based excess usage; the point
is that a nominal target is not itself an observed budget.

We evaluate Active RAG as budgeted decision-making rather than as a
single trigger-accuracy problem.  This framing exposes four failure modes that
unbudgeted comparisons can hide.  First,
\emph{utility heterogeneity}: retrieval may help, be neutral, or harm.  Second,
\emph{ranking failure}: a trigger score may not place beneficial cases above
neutral or harmful cases.  Third, \emph{calibration failure}: a threshold that
meets a budget on calibration data may miss it on held-out inputs.  Fourth,
\emph{cost accounting failure}: evidence-usage rate is not total deployment
cost, because uncertainty, probe-retrieval, and learned routers pay different
generation, retrieval, and prompt-token costs.

Our contribution is to make these operating points comparable.  We formulate
active retrieval as budgeted utility estimation, where utility is the marginal
correctness change from using retrieved evidence instead of a no-retrieval
answer.  We then operationalize an evaluation protocol that separates exact
ranking quality, deployable threshold transfer, conservative budget behavior,
retrieval harm, and heterogeneous trigger costs.  Finally, we instantiate the
protocol with representative trigger families---uncertainty scores, BM25
evidence scores, query-complexity triggers, and lightweight harm-aware utility
routers---and audit them across datasets, model families, random splits,
calibration sizes, manual and LLM correctness checks, and cost-aware
simulations.  The goal is not to crown a universal router, but to show how
future Active RAG systems can be compared at explicit, reproducible operating
points.

\section{Related Work}

\paragraph{Retrieval-augmented generation.}
RAG models combine generated text with evidence retrieved from an external
corpus \citep{lewis2020rag}.  This improves knowledge-intensive tasks but
introduces a cost-quality trade-off: retrieval expands the prompt and can add
irrelevant information.  Recent efficient and on-device RAG studies make the
same trade-off explicit at the system level, where retrieval quality, energy,
and resource use must be considered together
\citep{cheng2026toward,cheng2026energyefficientondeviceragmobile}.

\paragraph{Active and adaptive retrieval.}
FLARE retrieves during generation when low-confidence tokens indicate that
upcoming content needs evidence \citep{jiang2023flare}.  Self-RAG trains models
to emit reflection tokens that control retrieval and critique
\citep{asai2024selfrag}.  Adaptive-RAG selects among no-retrieval, single-step
retrieval, and multi-step retrieval using query complexity labels
\citep{jeong2024adaptiverag}.  DRAGIN estimates real-time information needs
during generation \citep{su2024dragin}, and UAR casts multiple active retrieval
criteria as plug-and-play classification tasks \citep{cheng2024uar}.  Recent
work also shows that uncertainty estimators remain competitive with more
complex adaptive pipelines \citep{moskvoretskii2025adaptive}.  Adjacent
agentic Text-to-SQL taxonomies frame LLM pipelines by inference-time autonomy
and feedback loops \citep{su2026agentic,zhao2026agentic}.  Related
document-routed RAG work frames retrieval selection as a
robustness--precision trade-off \citep{cheng2026resolvingrobustnessprecisiontradeofffinancial}.  Our focus is
orthogonal: regardless of the trigger family, thresholds should be calibrated
and evaluated under explicit usage budgets.

\paragraph{Calibration.}
Calibration asks whether model scores correspond to empirical outcomes
\citep{guo2017calibration}.  We use the term in a decision-theoretic sense:
trigger scores are not final answers, but decision variables that should be
thresholded on held-out data to meet a target usage rate.

\section{Budgeted Utility Frontiers}

The key distinction is between a \emph{score} and an \emph{operating point}.
A trigger score can be useful if it ranks high-utility retrieval decisions
above low-utility ones, but a deployed system also needs a threshold that
respects a budget on future inputs.  We therefore evaluate active retrieval
through two linked questions: how good is the ranking, and how reliable is the
calibrated threshold?

Let $q_i$ be a question, $y_i$ its gold answer, $a_i^0$ a no-retrieval answer,
and $a_i^R$ a retrieved-context answer.  The net utility of using retrieval for
that example is
\begin{equation}
  u_i = \mathrm{correct}(a_i^R, y_i) -
        \mathrm{correct}(a_i^0, y_i),
\end{equation}
where $u_i \in \{-1,0,+1\}$ corresponds to harmful, neutral, or beneficial
retrieval.  More generally, an active RAG policy $\pi$ chooses an action such as
no retrieval, single-shot evidence use, or a more expensive evidence strategy.
The deployment objective is
\begin{equation}
  \max_{\pi}\; \mathbb{E}[\mathrm{Acc}(a^{\pi(q)}, y)]
  \quad \mathrm{s.t.}\quad \mathbb{E}[C(\pi,q)] \le B .
\end{equation}
In our binary experiments, $\pi(q)\in\{0,R\}$ and the usage budget counts
examples whose final answer uses retrieved evidence.

\paragraph{Three frontiers.}
We distinguish three evaluation objects that are often conflated.  The
\emph{exact frontier} sorts held-out examples by a router score and retrieves
the top $\lfloor \rho n \rfloor$ examples at budget $\rho$; it is not a
deployable threshold, but diagnoses ranking quality under matched usage.  The
\emph{deployable frontier} chooses a threshold on a calibration split and
applies it unchanged to held-out inputs; it measures threshold transfer and
reports realized usage.  The \emph{conservative frontier} uses a stricter
threshold chosen with finite-sample slack; it measures how much accuracy is
lost when budget violations are discouraged.

\paragraph{Cost model.}
Our budget counts examples that use retrieved evidence in the final answer,
not a universal wall-clock or monetary cost.  Because trigger families pay
different pre-decision costs, we account for each policy type separately:
\begin{equation}
  \begin{aligned}
  C_m(q)=&\; C_m^{\mathrm{pre}}(q) \\
   &+ (1-\pi_m(q))C_m^{\mathrm{skip}}(q) \\
   &+ \pi_m(q)C_m^{\mathrm{use}}(q),
  \end{aligned}
\end{equation}
where $m$ indexes the router family.  Query-only routers have a cheap query
feature cost before the decision; skipped examples then pay for a no-retrieval
answer, while triggered examples pay for retrieval, retrieved-context prompt
tokens, and retrieved-context generation.  Uncertainty routers use the
no-retrieval generation itself as the pre-decision computation, so skipped
examples reuse that answer and triggered examples pay for retrieval plus a
second generation.  Retrieval-score routers pay a probe retrieval before the
decision; $C_{\mathrm{probe}}$ denotes this retrieval and can be reused by
triggered examples.  Full-feature BUR variants combine the no-retrieval
generation and probe retrieval before deciding.  This decomposition is narrower
than a full deployment latency model, but it makes the operating point explicit
and avoids conflating trigger computation with retrieved-context answering.
This cost-aware view is also related to sequential filtering analyses that
optimize decision pipelines under cost and selectivity assumptions
\citep{paranjape2026optimality}.

\paragraph{Reference utility routers.}
We evaluate several router families under this protocol rather than claiming a
single dominant trigger.  The lightweight Budgeted Utility Router (BUR) is a
reference learned router: it assigns a scalar score $s(q)$ that estimates
expected net retrieval utility.  The linear version trains a multinomial
logistic regression over $u_i\in\{-1,0,+1\}$ and scores examples as
$P(u_i=+1)-P(u_i=-1)$.  We also include a small gradient-boosted BUR variant
and feature ablations to test when utility modeling helps relative to simple
uncertainty, retrieval-score, and query-complexity baselines.

\paragraph{Budget-safe calibration.}
Nominal thresholds may violate the target budget on deployment inputs.  We
therefore evaluate a conservative quantile rule: instead of calibrating to
$\rho$, choose the threshold for
$\rho_{\mathrm{safe}}=\max(0,\rho-\epsilon)$, where
\begin{equation}
  \epsilon = \sqrt{\frac{\log(2/\delta)}{2n_{\mathrm{cal}}}} .
\end{equation}
Under the usual exchangeability assumption, this DKW-style slack motivates
population usage control, but we treat it empirically: conservative thresholds
should reduce held-out budget violations, possibly at the cost of accuracy.  We
report both nominal and conservative thresholds.

\section{Experimental Setup}

\paragraph{Datasets.}
We evaluate on three multi-hop QA benchmarks with candidate evidence
paragraphs: HotpotQA distractor \citep{yang2018hotpotqa}, 2WikiMultiHopQA
\citep{ho2020twowiki}, and MuSiQue \citep{trivedi2022musique}.  Our main
Qwen2.5-1.5B evaluation uses 2,000 validation examples per dataset with fixed
random seeds, filtering MuSiQue to answerable examples.  This controlled
setting isolates the trigger decision from large-scale indexing: retrieval is
BM25 over the candidate paragraphs for each question.  We interpret these main
results as controlled routing diagnostics rather than full open-domain RAG
performance.  To test whether the conclusions survive noisier evidence, we also
run HotpotQA global-candidate BM25 and dense-retrieval diagnostics: one index is
built over all 66,635 deduplicated validation candidate paragraphs, and each
query retrieves from that shared pool rather than from its own gold candidate
set.  This adds cross-example retrieval noise, but is still not a full
Wikipedia index.

\paragraph{Models and retrieval.}
Our main cross-dataset experiments use Qwen2.5-1.5B-Instruct
\citep{yang2024qwen25} on 2,000 examples per dataset.  We also run
SmolLM2-1.7B-Instruct on HotpotQA and Granite-3.1-2B-Instruct on HotpotQA plus
2Wiki as non-Qwen family checks
\citep{allal2025smollm2,ibm2024granite31}.  For each
example, the model first answers from memory.  We then retrieve the top five
BM25 paragraphs and ask the same model to answer using only those passages.
Answers are scored with exact match or token F1 above 0.80 after standard
normalization.

\paragraph{Triggers.}
We compare seven trigger scores spanning four families.  An
\emph{Adaptive-RAG-style} baseline uses
only the question text and cheap query-complexity features, approximating
query-level complexity routing rather than reproducing the original system.
We also evaluate \emph{mean entropy}, the average entropy over generated
no-retrieval tokens; \emph{minimum token confidence}, one minus the minimum
maximum-token probability; \emph{BM25 top score}; \emph{BM25 margin}, the
difference between the top two BM25 scores; \emph{BUR-linear}; and
\emph{BUR-GBM}.  BUR features include generation uncertainty, BM25 scores,
answer length, and question length.
We split each dataset evenly into calibration and held-out test sets,
stratified by utility class when possible.  Thresholds are selected only on the
calibration split.  The main tables use a CPU-only five-split evaluation over
budgets $\{10,25,50,75,90\}\%$, which reports mean accuracy, usage error,
violation rate, exact frontiers, and conservative frontiers.

\begin{table}[t]
\centering
\footnotesize
\setlength{\tabcolsep}{2.2pt}
\begin{tabular}{lrrrrrr}
\toprule
Dataset & No & Full & Ben. & Harm & Neut. & Net $u$ \\
\midrule
2Wiki & 21.3 & 28.5 & 15.9 & 8.8 & 75.3 & 7.1 \\
HotpotQA & 15.4 & 43.4 & 31.9 & 4.0 & 64.0 & 27.9 \\
MuSiQue & 2.2 & 11.9 & 11.2 & 1.5 & 87.3 & 9.7 \\
\bottomrule
\end{tabular}

\caption{Base rates for Qwen2.5-1.5B on 2,000 examples per dataset.  Values are
percentages.  No is no-retrieval accuracy; Full is always-retrieve accuracy;
Ben. is the fraction where retrieval fixes a no-retrieval error; harm is the
fraction where retrieval changes a correct answer to an incorrect one.}
\label{tab:base}
\end{table}

\balance
\section{Results}

We now ask the sequence of questions a deployer would have to answer before
trusting an Active RAG trigger.  Is retrieval utility actually heterogeneous?
Do scores rank useful retrieval decisions?  Does a calibrated threshold meet
the future budget?  Do the conclusions survive model and evidence changes?
And does evidence usage still tell the right story once trigger costs are
counted?

\paragraph{RQ1: Is retrieval utility heterogeneous?}
If retrieval only helped, Active RAG would mostly be a cost-saving problem:
retrieve whenever the budget allows.  Table~\ref{tab:base} shows that the
decision is more delicate because retrieval is not monotonic.
For Qwen2.5-1.5B, always retrieving improves HotpotQA accuracy from 15.4\% to
43.4\%, while retrieval benefit occurs on 31.9\% of examples.  The same model
sees a much smaller gain on 2Wiki, from 21.3\% to 28.5\%, because retrieval harm
is also higher at 8.8\%.  MuSiQue is the hardest setting in absolute accuracy:
no-retrieval answers are almost always wrong, but candidate retrieval raises
accuracy from 2.2\% to 11.9\%.

\begin{table}[b]
\centering
\scriptsize
\setlength{\tabcolsep}{1.6pt}
\begin{tabular}{lrrrrlr}
\toprule
Retrieval & Full & Ben. & Harm & Rand & Best@50 & Oracle \\
\midrule
Cand. BM25 & 43.6 & 33.4 & 3.4 & 28.8 & Adapt. 34.8 & 47.2 \\
Global BM25 & 34.6 & 26.8 & 5.8 & 24.4 & BUR-lin 30.0 & 40.8 \\
Global dense & 38.6 & 28.8 & 3.8 & 26.4 & Dense 30.8 & 42.8 \\
\bottomrule
\end{tabular}

\caption{HotpotQA retrieval-source diagnostic ($n=500$; no-RAG accuracy is
13.6\% for all rows).  Full is always-retrieve accuracy; Ben. and Harm are
utility rates; Rand and Oracle are exact-50 accuracies; Best@50 gives the best
router and exact-50 accuracy.  The harder global settings change both utility
and router ranking.}
\label{tab:global-bm25}
\end{table}

This heterogeneity also changes with the evidence source.
Table~\ref{tab:global-bm25} repeats the diagnostic under noisier retrieval.
Switching from candidate BM25 to the global HotpotQA BM25 index lowers
always-RAG accuracy from 43.6\% to 34.6\%, increases harm from
3.4\% to 5.8\%, and lowers the best exact-50 router from 34.8\% to 30.0\%.
A dense E5-small retriever recovers part of that gap, raising always-RAG to
38.6\% and lowering harm to 3.8\%, but still changes the best exact-50 router
and remains below the per-example candidate setting.  This is still not full
Wikipedia retrieval, but it demonstrates that evidence noise and retriever
family change both utility and router ranking.

\paragraph{RQ2: Do routers rank useful retrieval decisions consistently?}
Once retrieval has mixed utility, an active trigger is useful only if its score
orders examples well.  Table~\ref{tab:budget} shows that this ranking problem is
not solved by any single trigger family.  On
HotpotQA, BUR-linear reaches 33.8\% deployable accuracy at the 50\% target,
minimum token confidence reaches 33.3\%, entropy reaches 32.9\%, and the
query-only Adaptive-RAG-style baseline reaches 31.8\%.  These numbers are close
enough that a single operating point would invite over-interpretation.  Their
realized usage rates also differ, and their split-level violation rates range
from 40\% to 100\% when any overshoot counts, although tolerance-based
overshoot rates distinguish small finite-split excesses from larger misses.
Under an exact held-out 50\% top-$k$ diagnostic,
BUR-linear and minimum confidence both reach 32.7\%, entropy reaches 32.5\%,
and the oracle reaches 45.9\%.  This gap is itself a key reason to report
realized usage and exact diagnostics rather than only target budgets.
On 2Wiki, query-complexity and uncertainty scores are strongest around the same
target, while BM25 margin remains competitive and the confidence intervals
overlap.  On MuSiQue, the query-only Adaptive-RAG-style baseline is strongest
at the 50\% target, while entropy and BUR variants remain close under exact
usage.  These results support recent findings that uncertainty-based methods
are strong baselines \citep{moskvoretskii2025adaptive} and reinforce our main
claim: the protocol should not depend on one router family winning.

\begin{table*}[t]
\centering
\small
\setlength{\tabcolsep}{3pt}
\begin{tabular}{llrrrrr}
\toprule
Setting & Router & Deploy acc. & Usage & Err. & Split viol. & Exact acc. \\
\midrule
2Wiki / Qwen2.5-1.5B & Adapt. & 26.2$\pm$0.4 & 49.9 & 1.9 & 20.0 & 26.4$\pm$0.4 \\
 & Entropy & 24.9$\pm$0.7 & 49.9 & 1.2 & 60.0 & 25.0$\pm$0.9 \\
 & MinConf & 25.7$\pm$0.9 & 49.4 & 1.1 & 20.0 & 25.8$\pm$0.9 \\
 & BM25-top & 24.9$\pm$1.1 & 48.9 & 2.0 & 20.0 & 24.9$\pm$1.2 \\
 & BM25-mar & 25.3$\pm$0.8 & 49.8 & 1.3 & 40.0 & 25.4$\pm$0.7 \\
 & BUR-lin & 25.1$\pm$0.9 & 48.5 & 2.1 & 20.0 & 25.0$\pm$0.9 \\
 & BUR-GBM & 25.0$\pm$0.7 & 48.7 & 2.6 & 20.0 & 25.0$\pm$0.8 \\
\addlinespace
HotpotQA / Qwen2.5-1.5B & Adapt. & 31.8$\pm$1.4 & 49.3 & 2.3 & 40.0 & 31.9$\pm$1.3 \\
 & Entropy & 32.9$\pm$0.9 & 51.3 & 1.3 & 100.0 & 32.5$\pm$0.7 \\
 & MinConf & 33.3$\pm$1.6 & 51.5 & 2.9 & 60.0 & 32.7$\pm$0.9 \\
 & BM25-top & 31.5$\pm$1.2 & 49.1 & 2.3 & 40.0 & 31.8$\pm$0.8 \\
 & BM25-mar & 30.1$\pm$1.2 & 50.2 & 1.7 & 80.0 & 30.0$\pm$1.1 \\
 & BUR-lin & 33.8$\pm$1.1 & 51.6 & 2.1 & 60.0 & 32.7$\pm$0.9 \\
 & BUR-GBM & 33.3$\pm$1.4 & 51.3 & 2.2 & 80.0 & 32.8$\pm$0.9 \\
\addlinespace
MuSiQue / Qwen2.5-1.5B & Adapt. & 9.9$\pm$0.8 & 49.7 & 1.1 & 40.0 & 9.8$\pm$0.6 \\
 & Entropy & 8.5$\pm$0.6 & 51.3 & 2.8 & 60.0 & 8.4$\pm$0.2 \\
 & MinConf & 8.3$\pm$0.7 & 51.7 & 2.6 & 80.0 & 8.1$\pm$0.4 \\
 & BM25-top & 7.7$\pm$0.1 & 49.8 & 2.1 & 60.0 & 7.7$\pm$0.2 \\
 & BM25-mar & 8.0$\pm$0.3 & 50.7 & 2.1 & 60.0 & 7.9$\pm$0.2 \\
 & BUR-lin & 8.6$\pm$0.4 & 51.9 & 1.9 & 100.0 & 8.3$\pm$0.4 \\
 & BUR-GBM & 8.5$\pm$0.4 & 51.0 & 1.4 & 60.0 & 8.4$\pm$0.3 \\
\bottomrule
\end{tabular}

\caption{Five-split 50\% budget diagnostics for Qwen2.5-1.5B on 2,000 examples
per dataset.  Deploy acc. uses a calibration-selected threshold applied to
held-out examples.  Usage is realized held-out evidence usage, Err. is absolute
usage error, Split viol. is the fraction of random splits exceeding the target
by any amount, and Exact acc. enforces strict held-out top-$k$ usage.}
\label{tab:budget}
\end{table*}

\paragraph{RQ3: Do calibrated thresholds meet future budgets?}
The previous result separates ranking quality from a nominal target.  The next
question is whether a threshold chosen on calibration data behaves like the same
budget on held-out inputs.  Table~\ref{tab:budget} shows that this transfer is
imperfect even in the controlled setting.
In the five-split 2,000-example evaluation, nominal 50\% thresholds violate the
target in 28.6\% of selected router--split combinations for
2Wiki/Qwen2.5-1.5B, 65.7\% for HotpotQA/Qwen2.5-1.5B, and 65.7\% for
MuSiQue/Qwen2.5-1.5B.  The failure is
not that all routers retrieve too much; rather, target budgets, realized usage,
and accuracy must be reported together.

Budget-safe calibration illustrates the corresponding trade-off.  On
HotpotQA, BUR-linear's mean 50\% target usage drops from 51.6\% under nominal
thresholding to 46.5\% under the conservative rule, with accuracy moving from
33.8\% to 32.0\%.  The rule is conservative rather than free: for the
query-only Adaptive-RAG-style baseline on HotpotQA it reduces usage from 49.3\%
to 44.9\% and accuracy from 31.8\% to 30.6\%.  This illustrates the practical
trade-off between budget violation and answer quality.

To check whether automatic harm labels reflect real failures, we manually
annotate 60 retrieval-harm candidates.  Fifty-two are
valid harms.  Most valid harms are comparison errors or distractor-entity
switches, and annotators usually attribute them to generation over retrieved
evidence rather than to obviously wrong retrieval.  Thus retrieval harm is not
only an indexing problem: even relevant evidence can pull the generator toward
the wrong relation or entity.  Because all utility labels depend on automatic answer
correctness, we also run an expanded LLM-judge audit over 733 sampled rows
across datasets, model families, and retrieval settings, following the broader
use of LLMs as evaluators while treating their judgments as audit evidence
rather than ground truth \citep{li2025generation}.  The judge validates
93.3\% of automatic
benefit labels and 83.8\% of automatic harm labels, while near-threshold EM/F1
cases are much noisier at 40.0\% validity.  We therefore treat benefit/harm
rates as meaningful but report metric-threshold robustness rather than assuming
all automatic labels are exact, following broader concerns that benchmark
verdicts can shift with configuration choices, detectable-effect budgets, and
accuracy-only summaries
\citep{li2026safetyreproconfigurationconditionalrankinstability,
zhuang2026preregisteringdetectableeffectpairedmde,sun2026beyond,
wang2026timeseriesfoundationmodel,
han2026earlyearlyenoughdesigndependent,
wu2026classweightingversusconditioning}.

\paragraph{RQ4: Do rankings persist across model families?}
If thresholds are operating decisions, they should not be assumed to behave
identically across model families.  We therefore compare the same
evidence-usage frontier protocol across Qwen, SmolLM2, and Granite settings.

\begin{table*}[t]
\centering
\small
\setlength{\tabcolsep}{4pt}
\begin{tabular}{lrrrrrlrrr}
\toprule
Model & $n$ & No & Full & Benefit & Harm & Best@50 & Acc. & Rand & Oracle \\
\midrule
Qwen2.5-1.5B & 2000 & 15.4 & 43.4 & 31.9 & 4.0 & BUR-lin & 32.7 & 27.9 & 45.9 \\
SmolLM2 & 2000 & 9.8 & 30.6 & 23.5 & 2.6 & BUR-GBM & 23.4 & 20.0 & 33.1 \\
Granite-2B & 2000 & 13.6 & 33.2 & 25.8 & 6.2 & BUR-GBM & 28.7 & 23.7 & 39.7 \\
\bottomrule
\end{tabular}

\caption{HotpotQA model-family diagnostic with explicit sample sizes.  Full is
always-retrieve accuracy; Best, Rand, and Oracle are exact held-out top-$k$
accuracies at 50\% evidence usage.  Qwen2.5-1.5B, SmolLM2, and Granite use
clean 2,000-example runs.}
\label{tab:families}
\end{table*}

Table~\ref{tab:families} checks whether the phenomenon persists across model
families rather than only in the main Qwen setting.  The clean 2,000-example
SmolLM2 run has a lower no-retrieval baseline than the main Qwen run
(9.8\% vs. 15.4\%), but retrieval still
raises accuracy to 30.6\%, with 23.5\% beneficial cases and 2.6\% harmful
cases.  At exact 50\% usage, its best router reaches 23.4\% compared with a
20.0\% random policy; the oracle is 33.1\%.  Granite-2B gives the same
qualitative picture at clean scale: retrieval raises accuracy from 13.6\% to
33.2\%, but 6.2\% of examples are harmful; its best exact-50 router reaches
28.7\% compared with 23.7\% random and a 39.7\% oracle.  Qwen2.5-1.5B reaches
32.7\% at exact 50\% usage, compared with 27.9\% random and a 45.9\% oracle.
Across the clean 2,000-example rows, retrieval utility is non-monotonic, and
the best exact-50 router changes across models.  Thus the calibration protocol
transfers more cleanly than any individual router.

We also run a CPU-only multi-split calibration-size diagnostic after generation
is complete.  Across five random splits and
calibration sizes from 32 to 250, nominal 50\% thresholds violate the held-out
budget in roughly half of method--setting combinations, whereas budget-safe
thresholds nearly eliminate violations but retrieve conservatively.  We also
report tolerance-based overshoot rates, because a one-example excess and a
5-point excess should not be interpreted the same way.  This supports treating
budget behavior as a first-class deployment metric rather than a table
footnote.

\begin{figure}[t]
\centering
\includegraphics[width=\columnwidth]{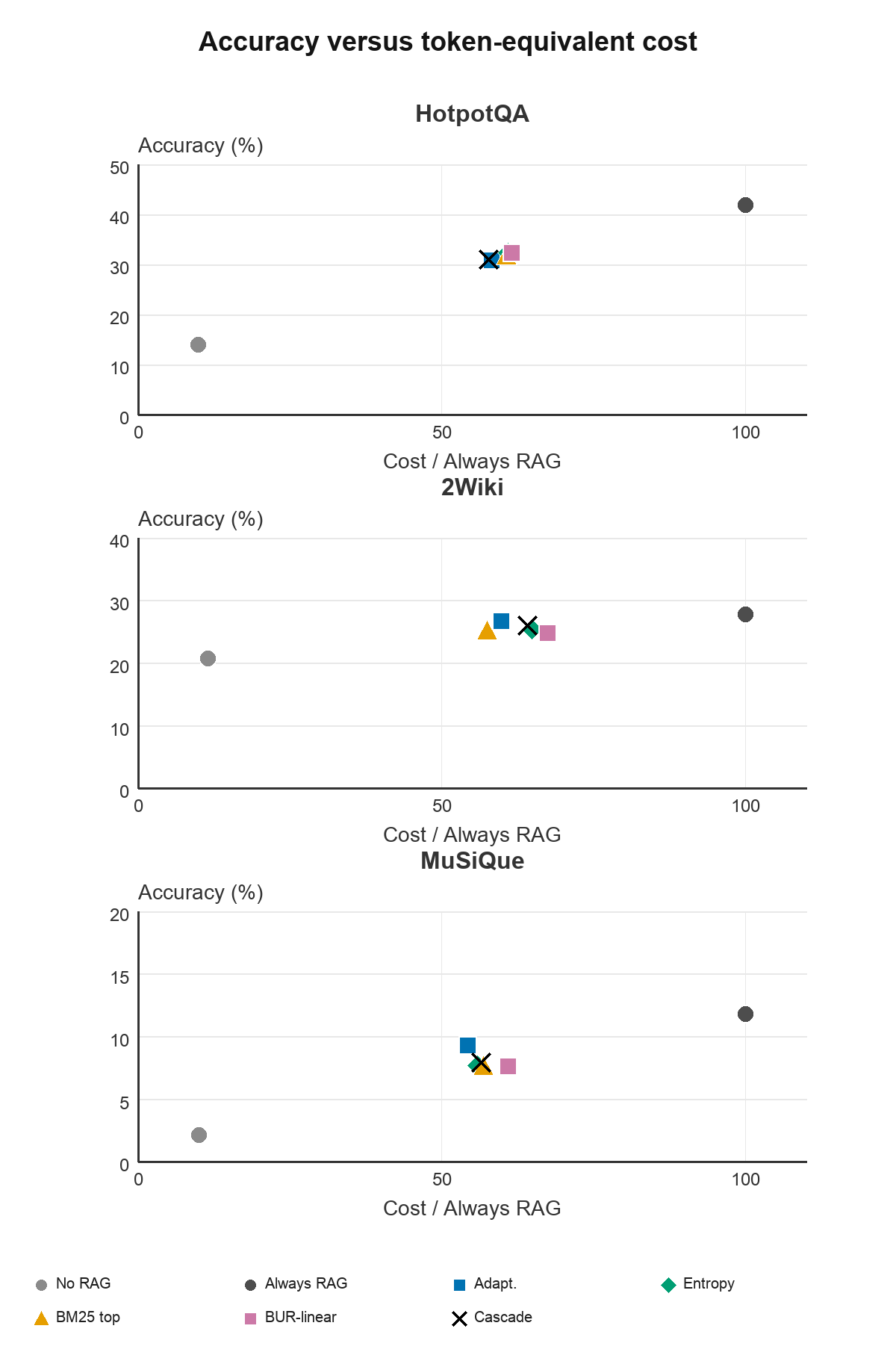}
\Description{Scatter plot comparing answer accuracy and normalized token-equivalent cost for active retrieval policies.}
\caption{Accuracy versus estimated token-equivalent cost normalized by
always-RAG.  Evidence-usage budgets do not fully determine deployment cost
because trigger families pay different pre-decision costs.}
\label{fig:cost-frontier}
\end{figure}

\paragraph{RQ5: How does cost accounting change the conclusion?}
Evidence usage is a convenient budget, but it is not the full cost of an active
policy.  Figure~\ref{fig:cost-frontier} simulates heterogeneous trigger costs with
tokenizer-aware prompt accounting.  The accounting separates query-only
routing, no-retrieval generation, probe retrieval, and retrieved-context
generation.  It shows why evidence usage alone is incomplete: on HotpotQA,
the clean 2,000-example cost simulation shows that BUR-linear can look
attractive under an evidence-usage budget while paying both no-RAG generation
and probe retrieval before many retrieved-context generations.  In contrast,
query-only routing avoids those trigger-side costs, and a three-stage cascade
reduces probe retrieval relative to BUR-linear.  The cascade improves the 2Wiki
accuracy-cost trade-off but is not uniformly best.  This supports reporting
cost decompositions rather than only retrieval rates.

\section{Discussion}

The central lesson is methodological.  Active RAG triggers are not merely
architectural switches; they are policies over uncertain future utility under a
budget.  Evaluating them only at one adaptive accuracy number can obscure
whether a method is genuinely better, better ranked but poorly calibrated, or
simply using retrieved evidence more often.  Exact frontiers and deployable
frontiers answer different questions: the former measure ranking quality at a
fixed usage rate, while the latter measure whether a calibrated threshold
transfers to held-out inputs.  Cost-aware accounting then asks whether the
chosen evidence-usage budget is the right proxy for deployment cost.  Reporting
all three makes the operating point explicit.

Our experiments are deliberately lightweight.  The main Qwen2.5-1.5B runs use
2,000 examples per dataset, and the SmolLM2 and Granite HotpotQA family checks
also use 2,000 examples; the global-candidate retrieval diagnostics
remain smaller.  The cost simulation also uses the clean
2,000-example Qwen2.5-1.5B setting, but remains a token-equivalent accounting
diagnostic rather than a deployment latency measurement.  We still rely mostly on
candidate paragraph retrieval rather than a full Wikipedia index, and on small
open instruction models.  These choices make the study reproducible on a
single consumer GPU, but they do not establish broad generality.  Future work
should add full-corpus retrieval, semi-structured retrieval with adaptive
fusion and reranking \citep{tao2026grasp}, 7B/API-scale instruction models,
multimodal recommendation-style retrieval with MLLM graph refinement
\citep{dang2025mllmrec}, and
long-form or long-context reasoning tasks where retrieval timing and context
compression jointly shape the budget \citep{gao2026dspc}.  It should also
evaluate model-side distillation, which has reduced inference requirements in
forecasting settings \citep{li2026distilling,li2025frequency}, as a
complementary efficiency axis alongside retrieval-side budget allocation.

\section{Conclusion}

We presented Active RAG as budgeted utility estimation rather than an
unbudgeted pipeline choice.  This perspective turns a vague question---whether
to retrieve---into three measurable questions: which examples should receive
retrieval, whether the calibrated threshold respects the future budget, and how
much computation the trigger itself consumes.  Across
HotpotQA, 2Wiki, and MuSiQue, retrieval benefit, retrieval harm, router
rankings, budget behavior, and cost rankings all vary substantially.  Future
Active RAG work should therefore report exact and deployable frontiers,
realized usage, threshold-transfer error, harm rates, and cost decompositions
alongside end-to-end task accuracy.

\section*{Limitations}

This study covers three multi-hop QA datasets and small open instruction
models, but still uses dataset-provided candidate paragraphs for the main
cross-dataset runs.  The HotpotQA global-candidate BM25 and dense diagnostics
add cross-example retrieval noise, but they are not true full-Wikipedia
open-domain indexes; absolute accuracies should therefore not be compared to
full QA systems.
Non-Qwen family checks remain limited: SmolLM2 is evaluated on HotpotQA, and
Granite is evaluated on HotpotQA plus 2Wiki.  Correctness is automatic and may
under-credit paraphrases, although our manual harm audit, expanded LLM-judge
correctness audit, and metric-threshold robustness diagnostic check the most
important error classes.
We do not yet evaluate long-form generation, multi-step retrieval,
memory-augmented agents \citep{liu2026memory}, API-scale models, or official
Self-RAG/FLARE reproductions inside the same frontier protocol; the comparisons
should therefore be read as representative trigger-family audits rather than
faithful rankings of named Active RAG systems.  Finally, our cost model separates trigger computation from
retrieved-context answering, but omits deployment-side latency, dollar, and
energy costs \citep{zhang2026carbon}.

\bibliographystyle{ACM-Reference-Format}
\bibliography{references}

\end{document}